\documentclass[10pt,twocolumn,letterpaper]{article}

\usepackage{cvpr}              
\definecolor{cvprblue}{rgb}{0.21,0.49,0.74}
\usepackage[pagebackref,breaklinks,colorlinks,allcolors=cvprblue]{hyperref}

\usepackage{multicol}
\usepackage{epsfig}
\usepackage{amsmath}
\usepackage{amssymb}
\usepackage{multirow}
\usepackage{booktabs}
\usepackage{color}
\usepackage{tabularx}
\usepackage{pifont}

\def\confName{CVPR}
\def\confYear{2026}

\title{SCL: Towards Domain Generalization via Single-Temporal Multimodal Contrastive Learning for Remote Sensing Change Detection}

\author{
Qiangang Du$^{1*}$\quad
Jinlong Peng$^{2*}$\quad
Xu Chen$^2$ \quad \\
Qingdong He$^2$ \quad
Liren He$^1$ \quad
Qiang Nie$^2$\quad
Mingmin Chi$^{1\dagger}$\quad \\
$^1$Fudan University, $^2$Tencent YouTu Lab, China\\
{\tt\small \{qgdu21,\,lrhe21\}@m.fudan.edu.cn,\quad mmchi@fudan.edu.cn,}\\
\, {\tt\small\{jeromepeng,\,cxxuchen,\,heqingdong\}@tencent.com,\quad qnie.cuhk@gmail.com}
}

\begin{document}
\maketitle

\def\thefootnote{}\footnotetext{$*$Equal contribution \quad $\dagger$Corresponding author}
\setcounter{footnote}{0}
\begin{abstract}
In recent years, change detection and anomaly detection models based on CNN and transformer have achieved remarkable success across various datasets based on paired data. However, most such methods exhibit limited cross-dataset generalization due to domain-specific designs and typically rely on large amounts of paired labeled data. In this paper, based on visual-language pre-training model, we introduce a \textbf{S}ingle-temporal multimodal \textbf{C}ontrastive \textbf{L}earning (SCL) foundation models for change detection without training on the target dataset. To further improve the model's ability to learn context of textual and visual information, we propose a \textbf{D}ynamic \textbf{T}ext-vision \textbf{C}ontext \textbf{O}ptimization (DTCO) module for prompt learning. Meanwhile, to address the data dependency issue of existing methods, we introduce a controllable generation and \textbf{S}ingle-temporal tr\textbf{AIN}ing strategy (SAIN). This allows us to train the model using a large number of existing single-temporal images without the need for paired label. Extensive experiments on various real-world change detection datasets demonstrate the superior performance and generalization of SCL, outperforming state-of-the-art methods under the evaluated settings. Code is available at \href{https://github.com/Kane-Du/scl-cd.git}{https://github.com/Kane-Du/scl-cd.git}.
\end{abstract}    
\section{Introduction}
Anomaly detection is a core direction in computer vision, aiming to identify patterns deviating from normal baselines and widely applied in industrial inspection. Notably, change detection, which focuses on identifying multi-temporal data differences, shares intrinsic connections with anomaly detection. As a specific form of temporal anomaly detection, change detection captures temporal deviations by comparing reference and target data, making up for traditional anomaly detection’s limitations in dynamic scenes.

\begin{figure}[tb]
    \centering
    \includegraphics[width=\linewidth]{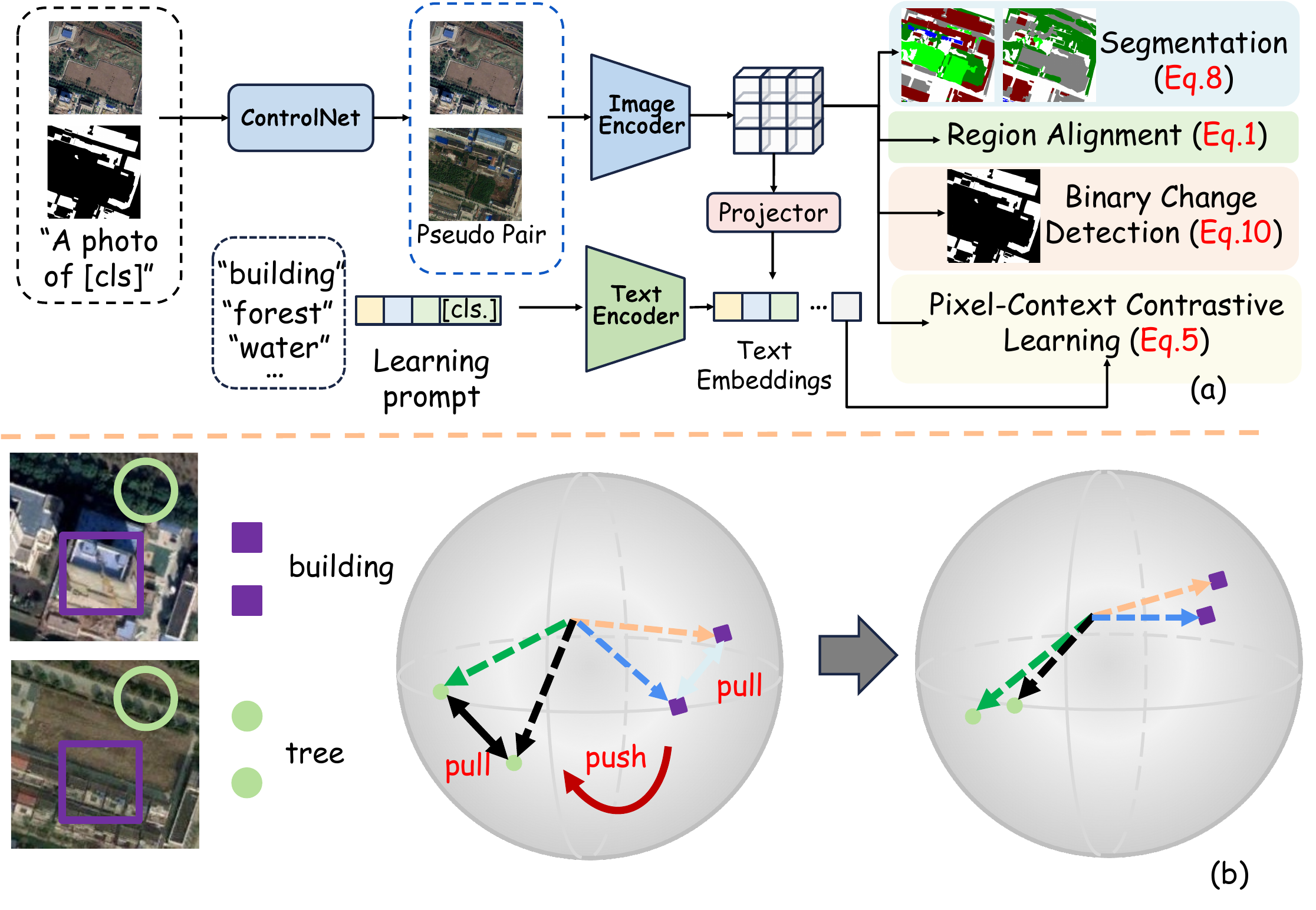}
    \caption{(a) SCL: Multimodal contrastive learning based on single-temporal for domain adaptation RSCD. (b) By employing the local patch-visual alignment and visual-context alignment in SCL, ``building''s and ``tree''s are clustered respectively and push away each other, while items within the same category are drawn closer together.}
    \vspace{-3mm}
    \label{fig:introduction}
\end{figure}
Current mainstream methods primarily focus on designing neural networks to improve prediction accuracy through optimization of network structure. These methods can be categorized into supervised change detection~\cite{BIT, chen2023sarasnet, arxiv:changer, ChangeFomer, noman2024elgcnet, noman2023scratchformer} and unsupervised change detection~\cite{Chen_2022_selfsup, Jong_2019_unsupCD, saha_2019_unsup,zhang2023diffucdunsupervised}. However, as focusing on addressing unique challenges within specific dataset, these methods are highly domain-specific. Learning on the domain-specific dataset causes the model to overfit to the particular data distributions, and leads the model to be oversensitive to changes. 

In response to this challenge, we draw inspiration from the excellent performance of multi-modal large-scale language models (MLLM) in zero-shot learning. We propose a \textbf{S}ingle-temporal multimodal \textbf{C}ontrastive \textbf{L}earning (SCL) foundation model for zero-shot and domain generalization change detection, as shown in Fig.~\ref{fig:introduction} (a). Our approach is built on CLIP~\cite{radford2021learning}. Driven by prior knowledge of visual-language pre-trained model, SCL achieves excellent result in the target domain without additional training, enabling the model to better cope with various challenges. Change detection method is expected to detect interest-changes by capturing complex data dependencies between bi-temporal image pairs. Since CLIP is image-text pairs contrast learning, it is inappropriate to directly adapt CLIP to dense change detection tasks. To address this problem, we adopt two alignments, local patch-visual alignment and visual-context alignment, to capture contextual information at the pixel level. The local patch-visual alignment is designed to minimize intra-class distances and maximize inter-class, as shown in Fig.~\ref{fig:introduction} (b). This innovation supports SCL's robustness to intra-class variations while preserving its sensitivity to different inter-class distinctions. Through these two alignment processes, SCL learns visual representations more effectively, improved clustering in feature space.

Most conventional approaches using expert-designed fixed prompts exhibit restricted flexibility in downstream adaptation. To some extent, it inhibits the generalization capacity of pre-trained models in downstream task applications, and compromises dynamic learning capabilities of visual encoder. Inspired by CoOp~\cite{zhou2022coop} and CoCoOp~\cite{zhou2022cocoop}, we introduce DTCO module of prompt learning. It enhances the ability of the model to capture interdependencies between visual and textual information, improving model cross-learning. This ability enables SCL to achieve excellent detection performance in the target domain without any training. Additionally, to alleviate the dependency on labeled data in change detection, we propose a controllable generation and single-temporal training strategy (SAIN) based on ControlNet~\cite{zhang2023controlnet}. This strategy allows us to generate a large number of change detection maps and improves the generalization capabilities of SCL by training on it. It makes SCL performs excellently on official change detection datasets without training on target datasets.

In summary, the main contributions of this paper are threefold:
\begin{itemize}
    \item To address the issue of poor domain generalization for change detection and anomaly detection, we propose a novel and practical contrastive learning foundation method called SCL. We capture data dependencies between image pairs through a local patch-visual alignment. Through visual-context alignment, text guides the change representation learning. Additionally, we enhance text-visual domain alignment via DTCO.
    \item To address the problem of change detection relying on labeled pair images, we propose a reliable method SAIN for constructing pairwise images from single-temporal image, which follows natural distribution.
    \item Our approach achieves strong quantitative results. Without additional fine-tuning on target datasets, SCL achieves superior performance against representative state-of-the-art methods across multiple change detection benchmarks as a unified model.
\end{itemize}
\section{Related Work}
\subsection{Bi-Temporal Training Change Detection}
Deep learning neural networks have demonstrated their excellence in change detection~\cite{arxiv:changer, wang2023apd, peng2021siamrcr}. Existing supervised change detection methods focus on foreground representation learning by special tricks, such as feature exchange~\cite{arxiv:changer, wang2023apd,noman2024elgcnet,chen2025damnetdomainadaptationnetwork}, attention mechanism~\cite{tgars:dessn, tgars:darn, 9259045Dasnet, peng2020chained,2024huangSEIFNet,2025tan_swcd,tian2025semanticsversusidentitydivideandconquer}, feature enhancement~\cite{tgars:dessn, MDENet,xu2025hsacnethierarchicalscaleawareconsistency}. These supervised methods heavily rely on the availability of extensive labeled datasets and require training a separate model for each dataset. Recently, unsupervised change detection methods reduce reliance on annotated data, yet their detection performance is often constrained in practical scenarios. Some of the work use GANs~\cite{Ren_2021_GAN,zheng2023changen} and Diffusion~\cite{zhang2023diffucdunsupervised,Jong_2019_unsupCD,wen2023gcdddpm} technology to reconstruct images as unsupervised change detection. They predict changes by mathematical methods such as principal component analysis (PCA)~\cite{kpca,saha_2019_unsup,noh2022unsupervised,dcae}. Xu~\cite{GMCD} et al. design proprietary pseudo-labeling generation structures to assist the above process. However,most supervised and unsupervised methods face the shared limitations: many tend to overfit to the training data distribution, leading to degraded cross-domain generalization performance.

Based on the powerful prior knowledge of CLIP~\cite{radford2021learning}, we propose SCL, which demonstrates strong generalization on RSCD datasets. SCL achieves excellent performance on change detection datasets without any additional training.
\subsection{Single-Temporal Supervised Change Detection}
Existing high-quality annotated change detection datasets~\cite{levir-cd, WHU, CDD2018Lebe} are small, but there are already a large number of remote sensing single-temporal annotated datasets~\cite{gupta2019xbd, loveda_wang_2021}. ChangeStar~\cite{Zheng_2021_ICCV_changestar} first proposed a single-temporal change detection training strategy STAR which forms the pseudo-pair by randomly matching within the training batch. However, uncontrolled random matching leads to a significant overlap of buildings and other changes in the image pairs that contradict objective reality. It is difficult to perform shape analysis. Moreover, since the dataset images are collected at the same time, it cannot simulate the changes caused by seasons and other factors in natural scenes. Changen~\cite{zheng2023changen} generates pseudo-image pairs based GAN~\cite{mirza2014conditional}, but its controllability and flexibility are limited. In recent years, MLLM has also shined brilliantly in the field of change detection~\cite{zheng2024anychange, DONG202453ChangeCLIP}.

In this paper, we propose a single-temporal training strategy based on text-to-image generation model, with text-image mask controllable generation. By generating abundant trainable data through directed category editing, we mitigate data dependency. Through the generated data, we improve the model's generalization ability significantly.
\begin{figure*}[tb]
    \centering
    \includegraphics[width=0.8\linewidth]{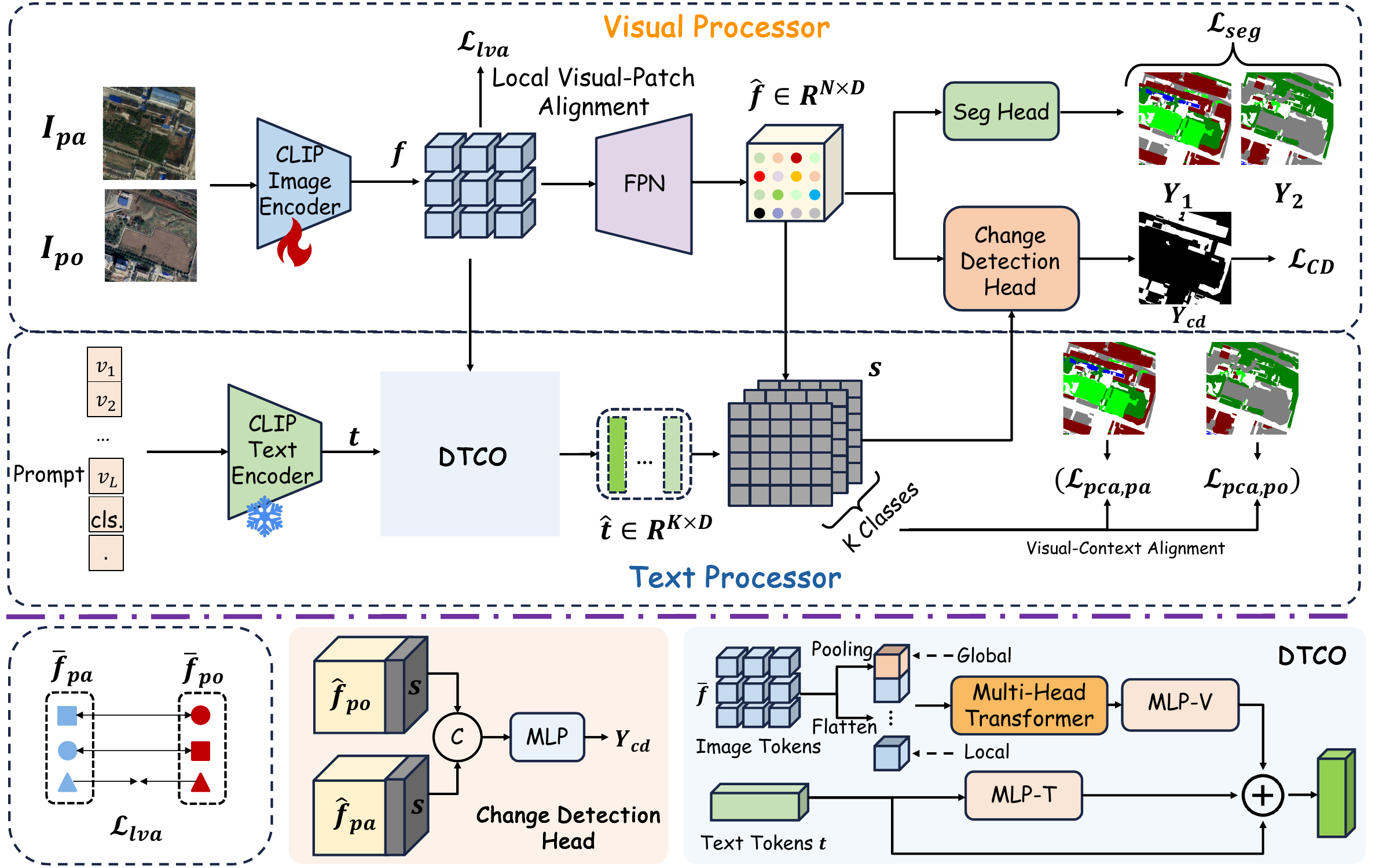}
    \caption{Illustration of SCL. SCL can be divided into two parts: visual processor and text processor. At the bottom are specific structure diagrams of local visual-patch alignment, change detection head and DTCO. Firstly SCL extracts the region visual-patch features $\bar{f}$ and gets the dense visual features $\hat{f}$ by FPN. Then a K-way classification is used for semantic segmentation. And the local visual-patch alignment  loss $\mathcal{L}_{lva}$ is used to alleviate the lack of large-scale pre-trained model on region feature learning. Then the text encoder extracts text embeddings and Fine-tuning CLIP for change detection of generalisability through visual-context alignment loss $\mathcal{L}_{pca}$.}
    \label{fig:framwork}
\end{figure*}
\subsection{Vision-Language Models}
Computer vision has developed rapidly, driven by the pre-trained + finetuning paradigm. Pre-trained models such as Contrastive Language-Image Pre-training (CLIP)~\cite{radford2021learning} and Imagen~\cite{Imagen} are even more impressive for ``text+vision'' multimodal large language models. But these works are based on the whole image classification. And the prompts provided by specialists are fixed. DenseCLIP~\cite{rao2021denseclip} and MaskCLIP~\cite{zhou2022maskclip} propose CLIP-based methods to transfer CLIP to downstream dense prediction. CoOp~\cite{zhou2022coop} and CoCoOp~\cite{zhou2022cocoop} validate the effect of prompt engineering on pre-trained models and they adapt visual-language pre-trained models like CLIP for downstream image recognition. 

 In this paper, we propose a paradigm of pixel-wise text-vision contrastive learning and visual region contrastive learning, demonstrating superiority in dense pixel recognition and classification for RSCD. In addition, we propose a method of dynamic text-context optimization (DTCO), which ensures both the utilization of powerful prior knowledge and the generalization of the pre-training model. 
\section{Method}
The overall framework of SCL is shown in Fig.~\ref{fig:framwork}. SCL consists of three parts: (i) local patch-visual alignment; (ii) visual-context alignment; (iii) semantic segmentation and change detection. We feed the bi-temporal images, generated from single-temporal images, $I_{po}$ and $I_{pa}$ into CLIP image encoder to obtain patch-visual embeddings $f_{po}$ and $f_{pa}$. And the text embeddings $\bar{t}$ is obtained from the learnable prompt $t$ via CLIP text encoder.

\subsection{Contrast Learning and Multi-Task Detection}\label{sec:alignment}
\subsubsection{Local Patch-Visual Alignment.} \label{sec:loc-algin}
There is a large amount of pseudo-change noise between the bi-temporal images. Therefore a satisfactory image encoder should capture important visual information and be more robust to variations. As shown in Fig.~\ref{fig:introduction} (b), In feature space, our goal is to bring the features of the same class closer and push the features of different classes farther apart. This ensures that the feature representations of targets are intra-class compact and inter-class separable.

Conversely, the features of different categories should be pushed apart. To handle this, we employ a contrastive loss on the local region features $\bar{f}_{po}$ and $\bar{f}_{pa}$, shown as in Fig.~\ref{fig:framwork}. The $\bar{f}_{po}$ and $\bar{f}_{pa}$ are the normalized embeddings of $f_{po}$ and $f_{pa}$. Let $y_{po}$ and $y_{pa}$ denote the ground-truth of post-temporal and past-temporal image, and $\tau$ represents the temperature parameter to control the concentration of the similarity distribution. Based on this, we can obtain the similarity label by bitwise AND operation:
$\bar{y}_i=y_{po,i}\ \&\ y_{pa,i}$. The cosine similarity distance between the $i$-th pixel features from the past and post-temporal images is given by $d_i=\left \langle \bar f_{i,pa}, \bar f_{i,po}\right \rangle$. 
We define the local patch-visual alignment loss $\mathcal{L}_{lva}$ mathematically as Eq.~\ref{pix-loss}.
\begin{equation}\label{pix-loss}
    \mathcal{L}_{lva} =  -\frac{1}{N}\sum_i^N \bar{y}\log d_i + (1-\bar{y}_i)\log(1-d_i)
\end{equation}
where $N$ is the total number of pixels.

\subsubsection{Visual-Context Alignment.} \label{sec:vxt-align}
With the pixel-wise image representations $\hat{f}_{po}, \hat{f}_{pa} \in R^{N\times D}$ and text embeddings $\bar{t} \in R^{K\times D}$, we aim for the model to leverage prior knowledge of CLIP to learn the information dependency between text and vision through pixel-wise contrastive learning. $N$ is the pixel number, $D$ is the number of embedding dimensions, and $K$ is the class number. Our purpose is to align textual and pixel-level visual information and facilitate effective cross-modal understanding.
The learnable prompt $t_k$ is defined as:
\begin{equation}\label{eq:t_k}
    t_k = TE([v_1, v_2...v_{l-1}, c_i])
\end{equation}
where $\{v_1, v_2...v_{l-1}\}$ denotes the learnable context vectors, $c_i$ is the word embedding of the $k$-th class word name, $l$ denotes the text context length, and $TE$ is the text encoder of CLIP.
The category prediction of each pixel pair $f_i \in \{\hat{f}_{i,po}, \hat{f}_{i,pa}\}$ is achieved through classification:
\begin{equation}
    \hat{k}_{f_i} = \mathop{\arg\min}\limits_{(k)}\{\left \langle f_i, t_{k}\right \rangle\}_{k=1}^{K}
\end{equation}
We define the probability of pixel $i$ over the class:
\begin{equation}
    p(k|f_i) = \frac{\exp(-s_{i,k})}{\sum_{k'=1}^{K}exp(-s_{i, k'})}
\end{equation}
where $s_{i,k}=\left \langle f_i, t_{k}\right \rangle$ denotes the cosine similarity of $f_i$ and text embeddings $t_k$ of class $k$. Given the ground-truth $y$ of bi-temporal image $I_{pa}, I_{po}$, i.e., $y_i\in\{1,\dots,K\}$, the pixel-context alignment loss can be computed as:
\begin{equation}
    \mathop{\mathcal{L}_{pca, j}} =-\frac{1}{N}\sum_{i=1}^{N}y_{i}\log{\frac{exp(-s_{i,k})}{exp(-s_{i,k})+\sum_{k'\ne k_i}exp(-s_{i, k'})}}
\end{equation}
where $j\in \{pa, po\}$.

\subsubsection{Semantic Segmentation and Change Detection.}\label{sec:detect}
From an intuitive perspective, change detection can be seen as a parallel task to semantic segmentation. Existing pre-training models are unaware of the region-level and pixel-level image representations with the text tokens. To address this, we employ a Feature Pyramid Network (FPN) to obtain both region and pixel level representations. The FPN upsamples the features $f_{po}$ and $f_{pa}$ of the bi-temporal patches from 1/32 ($f$) to 1/4 ($\hat{f}$) of the original image size, shown as in Fig.~\ref{fig:framwork}. We represent this densely upsampled pixel-wise features as $\hat{f}_{po}$ and $\hat{f}_{pa}$. Then we use a lightweight classification network to predict the semantic segmentation results for the bi-temporal images. Each pixel embedding $\hat{f}_i\in R^{D}$ is fed into a $K$-way classification, where $i$ is the pixel index. Assuming the pixel $\hat{f}_i$ belongs to class $k$, the probability of $f_i$ being classified as class $k$ is computed as: 
\begin{equation}
    p(k|f_i) = \frac{\exp(w_{k}^{T}\hat{f}_i)}{\sum_{k=1}^{K}exp(w_{k}^{T}\hat{f}_i)}
\end{equation}
where $\hat{f_i}\in \{\hat{f}_{i,po}, \hat{f}_{i,pa}\}$ denotes the feature of $i$-th pixel, $W$ denotes the learnable parameter of pixel-wise Seg Head in Fig.~\ref{fig:framwork}, $W=[w_1,...,w_k]\in R^{K\times D}$, $w_k \in R^D$ .
The loss function is defined as follows:
\begin{align}\label{loss-seg}
    \mathcal{L}_{seg} &= -\log p(k|\hat{f}_i) = -\sum_{i=1}^{N} (y_{i,po} log p_{i,po} - y_{i,pa} log p_{i,pa})
\end{align}

With the visual features $\hat{f}_{po}, \hat{f}_{pa}$ for dense pixels in the bi-temporal images, and the contrastive scoring map $s$ between visual features and context embeddings, we employ a lightweight detection head to detect object changes. The visual information includes visual features such as object shapes, while the scoring map contains pixel-level class information. Therefore, we concatenate the scoring map with the visual feature map for the detection head. 
\begin{equation}
    \boldsymbol{p_k} = softmax(MLP([\hat{f}_{po}, s_{po}, \hat{f}_{pa}, s_{pa}]) )
\end{equation}
where $p_k\in[0,1]$ denotes the predicted probability for changed objects, MLP is the detection head.
This combined input allows the detection head to leverage both the visual and class information for accurate object change detection. 
With the change detection ground-truth $\hat{y} \in\{0,1\}$, we adopt binary cross-entropy loss for change detection loss:
\begin{equation}
    \mathcal{L}_{cd}(\boldsymbol{p_k}, \hat{y}) = -\hat{y}\log(\boldsymbol{p_k}) + (1-\hat{y}) \log(1-\boldsymbol{p_k})
\end{equation}
where $\hat{y}\in\{0,1\}^{H\times W}$ specifies the ground-truth, $0$ is unchanged object and $1$ is changed object.

\subsubsection{Multi Supervision.}
The overall loss function includes the losses of four components, as following:
\begin{equation}\label{eq:loss-overall}
    \mathcal{L} = \mathcal{L}_{seg} + \mathcal{L}_{cd} + \alpha\mathcal{L}_{lva} + \beta(\mathcal{L}_{pca,po}+\mathcal{L}_{pca,pa})
\end{equation}

Since the fundamental tasks are change detection and instance segmentation, the main components of the loss function are the semantic segmentation loss $\mathcal{L}_{seg}$ and the change detection loss $\mathcal{L}_{cd}$. The local patch-visual alignment loss $\mathcal{L}_{lva}$ and visual-context alignment loss $\mathcal{L}_{pca,po}, \mathcal{L}_{pca,pa}$ serve as auxiliary losses in this context. Therefore, we regulate the auxiliary losses weights by $\alpha=0.1, \beta=0.1$, and the main loss weight is always set to $1$.
\subsection{Dynamic Text-Vision Context Optimization}\label{sec:dtco}
DenseCLIP~\cite{rao2021denseclip} and CoCoOp~\cite{zhou2022cocoop} utilize lightweight networks to learn the visual-textual context representation. They then perform contrastive learning by directly adding these new contextual representations to visual information. This method limits the model's flexibility and lacks the ability to capture dynamic interactions between visual contextual information and textual information. It undermines the generalisability of CLIP and causes the model to overfit to that particular context. To address this, we propose a dynamic text context optimization (DTCO) to better adapt the visual features to context.

Initially, we utilize a transformer network based on cross-attention to map visual features into the textual feature space. The multi-head cross-attention mechanism establishes text-visual connections for each pixel. It allows the model to capture the interdependencies and correlations between visual and textual information, enabling a more comprehensive understanding of the visual context. Then, we employ two small networks, known as adapters, to separately map the visual information to the textual space and the textual information to the visual space. The adapters determine the relevance of visual and textual information in different contexts. Finally, we use a learnable dynamic weight to combine the outputs of these adapters, resulting in a new contextual embedding. This dynamic weight allows the model to adaptively adjust the importance of visual and textual information, enhancing the fusion process.

Based on the aforementioned process, we can define the contextual representation as follows:
\begin{equation}
    z =Transformer([f,g], t_k), g=Pooling(f)
\end{equation}
\begin{equation}
    \hat{t}_k = \varphi(z)*t_k + \varsigma(t_k)*z + \sigma t_k
\end{equation}
where $\varphi, \varsigma$ is a lightweight network (MLP-V and MLP-T in Fig~\ref{fig:framwork}), $\sigma$ is the weight vector and $\sigma \in R^{L}, \hat{t}_k\in R^{K\times L}$, $t_k$ is calculated in Eq.~\ref{eq:t_k}.
By dynamically optimizing the textual context and focusing on relevant textual features for pixel-wise visual feature, the model effectively captures the mutual dependencies and associations between visual and textual information. 
\subsection{Controllable Generation and Single-Temporal Training Strategy}\label{sec:strategy}
\begin{figure}[tb]
    \centering
    \includegraphics[width=\linewidth]{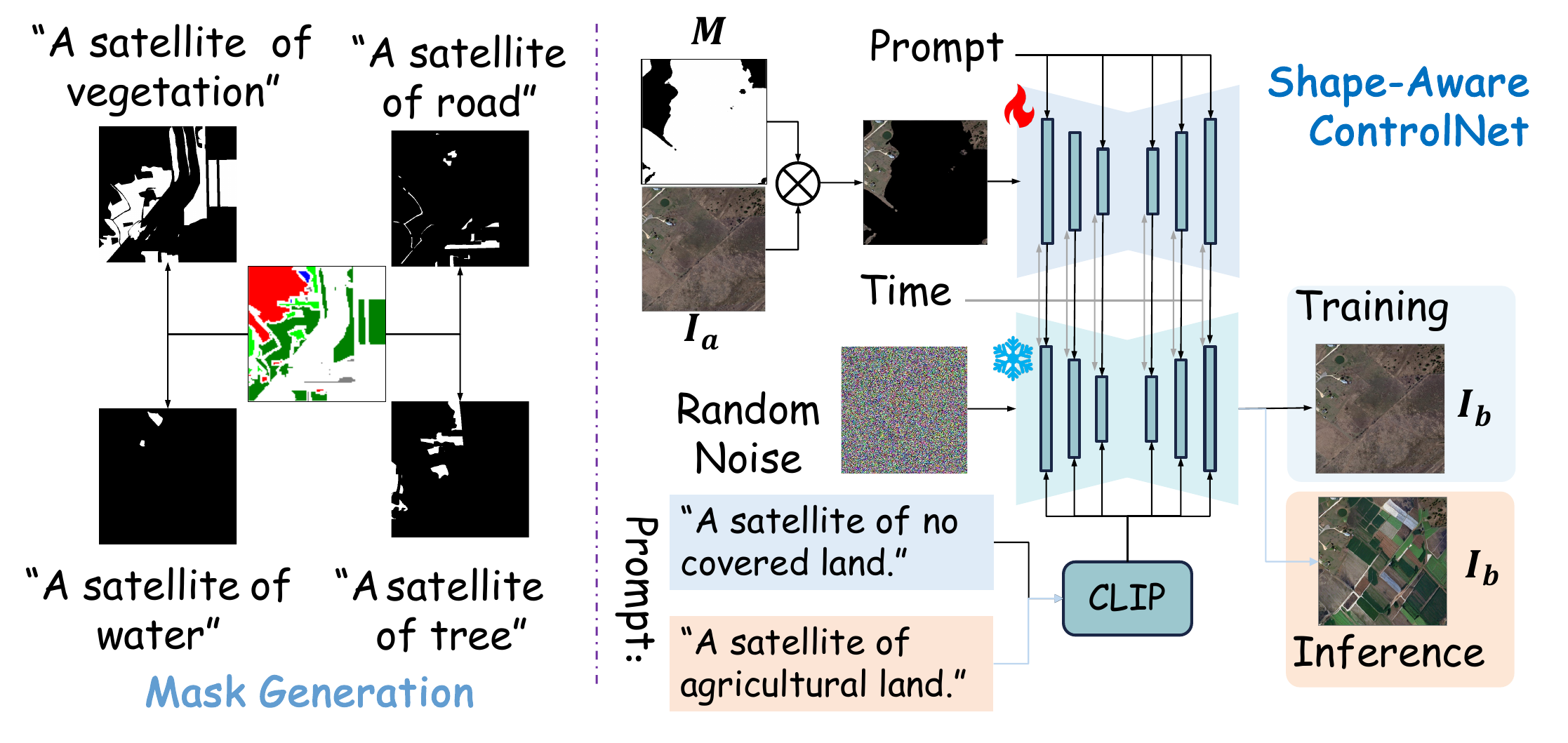}
    \vspace{-1mm}
    \caption{Framework of SAIN. We separate each class of objects and generate shape masks by mask generation. During the training phase, we train the ControlNet~\cite{zhang2023controlnet} by inpainting learning. In inference phase, we introduce deliberate category-based changes to each object in the image except buildings.}
    \vspace{-2mm}
    \label{fig:strategy}
\end{figure}
Acquiring high-quality instance annotations for bi-temporal image pairs comes at a considerable cost. Revisiting the essence of temporal image pairs for change detection, it is evident that pixels at the same location in these pairs represent two distinct states of the same geographical area at different times. These states may manifest as different physical conditions of the same object over time or as changes induced between different categories of objects at the same location. Fortunately, a large number of single-temporal remote sensing instance segmentation images are already available, which are highly relevant to change detection tasks. ChangeStar~\cite{Zheng_2021_ICCV_changestar} proposes a method of constructing image pairs by STAR~\cite{Zheng_2021_ICCV_changestar}. However there are some shortcomings, the generated data deviate from practical significance, and violate the true changes observed in the natural world. The establishment of correspondence between these diverse states is fundamental to train object change detection models based on single-temporal images.

To address the aforementioned issue, we propose SAIN to generate pseudo-image pairs for feeding enough data to SCL, as shown in Fig.~\ref{fig:strategy}. Controlled image editing provides a promising approach for addressing this issue. By selectively editing specific regions of an image, one can obtain pairs of images closely resembling each other, thus simulating different states over time. 

In this paper, we implement this approach based on ControlNet~\cite{zhang2023controlnet}, denoted as $\mathcal{G}$. We begin by defining an image $\mathcal{I}_a = \{x_1, x_2, \dots, x_i, \dots, x_N\}\in R^{N\times 3}$, composed of $N$ objects belonging to $K$ categories, where $x_i\in\{C_1, \dots, C_K\}$, each with its segmentation annotation map $g\in{1,...,K}$. 
We define the process to generate pseudo-image as shown in Eq.~\ref{image-gen}.
\begin{equation}\label{image-gen}
    I_b = \mathcal{G}(\mathcal{I}_a, M, P_t, n)
\end{equation}
where $\mathcal{I}_a$ is the real image, $M$ is a binary mask map to control the edited area, $P_t$ is the prompt which specifies the category information of the object in the edited area, and $n$ is the random noise.

During the training phase, we finetune the ControlNet~\cite{zhang2023controlnet} by image inpainting approach to recognize objects in each category. In the inference phase, changes between categories or within the same category are generated through predefined priors. We can obtain the generated change map as following:
\begin{equation}
    \hat{I}_b = \mathcal{G}(\mathcal{I}_a, M, \hat{P_t}, n)
\end{equation}
where $\hat{P_t}$ is the prompt containing the generation information of changes in the edited area.

SAIN provides a cost-effective and feasible solution to address the challenge of acquiring bi-temporal annotated data for change detection. Our proposed method generates realistic and informative pseudo-image pairs and substantially improves performance of change detection models.
\section{Experiment}
\subsection{Datasets}
Two high spatial resolution datasets (xView2~\cite{gupta2019xbd}, SECOND~\cite{yang2020asymmetric}) and the SAIN generated dataset are used to train models. Two RSCD datasets (LEVIR-CD~\cite{levir-cd} and WHU-CD~\cite{WHU}) are used to evaluate the performance.

\begin{table*}[tb]
\setlength{\tabcolsep}{4pt}
\large
\centering
\resizebox{\textwidth}{!}{
    \begin{tabular}{c|ccc|ccc|ccc|ccc|ccc|ccc}
    \toprule
    \multirow{3}*{Model} & \multicolumn{6}{|c|}{xView2} & \multicolumn{6}{|c}{SECOND} &  \multicolumn{6}{|c}{SAIN~\ref{sec:strategy}}\\
    \cline{2-19}
    ~ & \multicolumn{3}{|c|}{LEVIR-CD} & \multicolumn{3}{|c|}{WHU-CD} & \multicolumn{3}{|c|}{LEVIR-CD} & \multicolumn{3}{|c}{WHU-CD} & \multicolumn{3}{|c|}{LEVIR-CD} & \multicolumn{3}{|c}{WHU-CD} \\ 
    \cline{2-19}
    ~ & F1 & IoU & OA & F1 & IoU & OA & F1 & IoU & OA & F1 & IoU & OA  & F1 & IoU & OA  & F1 & IoU & OA \\ 
    \hline
    BiT~\cite{BIT}  &   66.12   &   49.35   & 97.05 & 62.47 & 45.42&96.90& 70.23   &   53.23   & 97.42  & 66.94 & 50.31 & 97.35 & 80.00 & 66.67 & 98.20  & 69.74 & 53.55 & 96.70 \\
    ChangeFormer~\cite{ChangeFomer}  &   66.91   &  50.27    & 96.56 & 59.31& 42.16& 96.01 & 67.47& 50.92 & 94.12& 65.53 & 48.73&93.86 & 72.55& 56.93 & 97.85 & 68.08 & 51.61& 97.31\\
    Changer~\cite{arxiv:changer}  &  62.26   &    45.20   & 96.74 & 37.96& 23.43& 94.05 & 57.13& 39.99 & 94.86 & 62.30 & 45.24& 95.71& 68.83 & 52.47 & 96.95 & 65.59 & 48.79 & 96.76 \\
    SARAS-Net~\cite{chen2023sarasnet}  &  45.59    &   29.53    & 92.09 &32.41 & 19.34  & 84.13 & 40.10& 25.08 & 89.93& 16.59 &9.05 &57.12 & 52.28 & 35.39 & 87.49 & 34.21 & 20.63& 87.49  \\
    TinyCD~\cite{codegoni2022tinycd}  &  56.65   &   39.52    & 94.97 & 29.52 &  17.32 & 88.18 &67.43 & 50.87 & 96.07& 60.32 & 43.18 & 95.55 & 68.21 & 51.76 & 95.83 & 74.28 & 59.08 & 97.62\\
    USSFC-Net~\cite{ussfc}  &  49.32   &   32.73    & 94.49 & 44.81 & 28.88& 93.27 & 70.34 & 54.25& 97.37 &50.83 & 34.08 & 95.22 & 72.63 &  57.01& 97.23& 63.52 & 46.54 & 97.02 \\
    \hline
    ChangeCLIP~\cite{DONG202453ChangeCLIP}  &   73.80   &   58.48    & 97.74 &56.48 &39.35 & 96.79 & 69.37 & 53.10 & 97.44 & 67.53 & 50.98 &  97.34 & 68.89 & 81.58 & 98.13 & 57.29& 72.85 &  97.68 \\
    ChangeStar~\cite{Zheng_2021_ICCV_changestar}  &   79.31   &   65.71    & 98.06 & 73.59& 58.22& 97.43 & 54.67 & 37.61 & 96.72& 50.76 & 34.01 & 92.84 & 70.53 & 62.93 & 98.07 & 75.96& 55.40 & 97.41  \\
    SCL(Ours)    &    \textbf{89.02}   &    \textbf{70.74}   &   \textbf{98.29}  & \textbf{78.64}& \textbf{61.64} & \textbf{97.92} & \textbf{73.56}& \textbf{64.44} &\textbf{98.11} & \textbf{78.36} & \textbf{64.41}& \textbf{98.06}& \textbf{85.19} & \textbf{74.20} & \textbf{98.59} & \textbf{79.92} & \textbf{64.67} & \textbf{98.61} \\
    \bottomrule
    \end{tabular}
    }
    \caption{Zero-shot change detection on LEVIR-CD~\cite{levir-cd} and WHU-CD~\cite{WHU} datasets. The model is trained on xView2~\cite{gupta2019xbd}, SECOND~\cite{yang2020asymmetric} and SAIN~\ref{sec:strategy} generated. }
    \label{tab:benchmarks}
\end{table*}

LEVIR-CD~\cite{levir-cd} dataset contains 637 bi-temporal remote sensing image pairs with size of $1024^2$. LEVIR-CD is officially split into three parts, including train/val/test sets of samples 445/64/128. We merge all sets as the validation.

WHU building change detection~\cite{WHU} dataset contains only one image pair with size of 32507 × 15354. We crop it into size of $512^2$ and obtain 1920 pairs for evaluation. 
\subsection{Metrics and Implementation Details}
\begin{figure*}[tb]
    \centering
\includegraphics[width=0.8\linewidth]{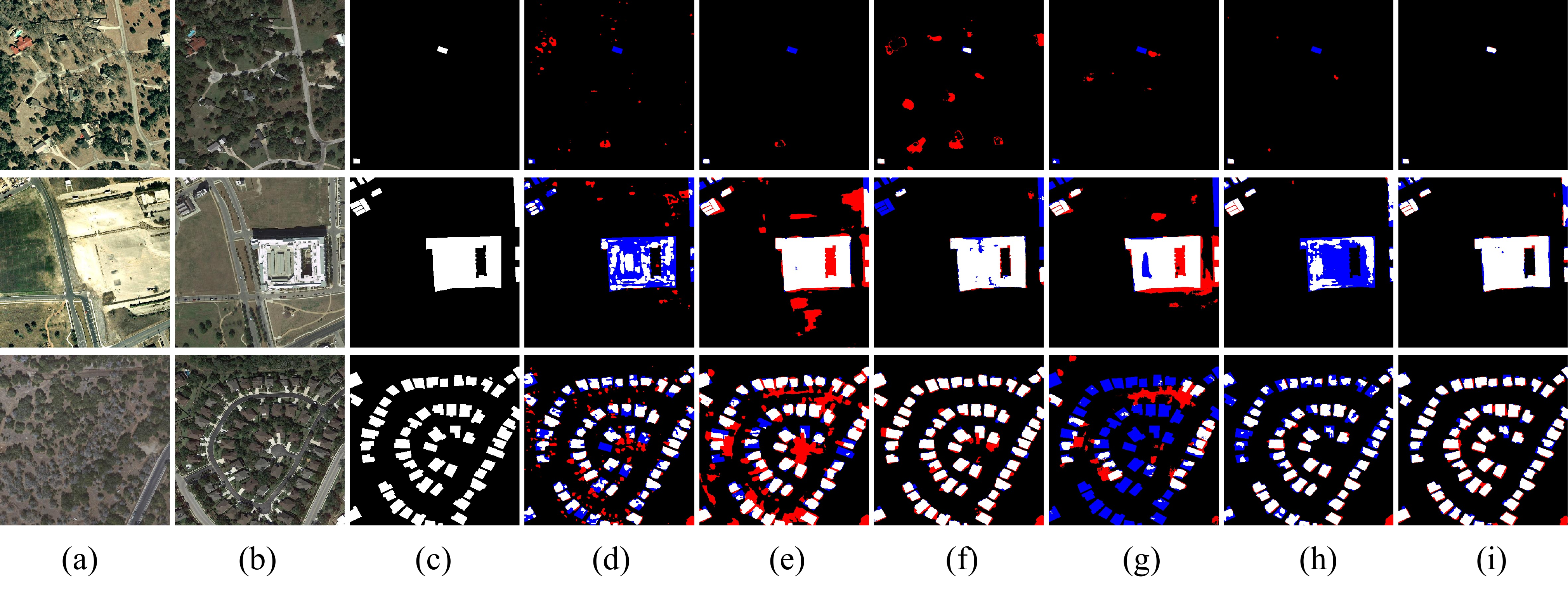}
\vspace{-1em}
    \caption{Visualization analysis for SCL with benchmarks. The \textcolor{red}{red} regions indicate False Positives (\textcolor{red}{FP}), while the \textcolor{blue}{blue} regions denote False Negatives (\textcolor{blue}{FN}). Each column represents the respective performance of (a) Past-temporal, (b) post-temporal, (c) groundtruth, (d) Changer~\cite{arxiv:changer}, (e) SARAS~\cite{chen2023sarasnet}, (f) Tiny-CD~\cite{codegoni2022tinycd}, (g) USSFC~\cite{ussfc}, (h) ChangeStar~\cite{Zheng_2021_ICCV_changestar}, (i) SCL.}
    \label{fig:badcase}
    \vspace{-1mm}
\end{figure*}
\textbf{Evaluation metric.}
In this context, we adopt F1-score, IoU and OA as the primary focus centers. Our method is implemented on CLIP~\cite{radford2021learning}, utilizing ViT-B~\cite{dosovitskiy2020vit} as the backbone. The parameters of the text encoder are frozen. The image encoder is fine-tuned, with the gradient weight parameters set to 1e-5. We use standard data-augmentation techniques and image pair swapping operations. We adopt AdamW as parameter optimizer, with the learning rate schedule following the polynomial annealing policy. LEVIR-CD and WHU-CD, merged all the official sets, are used to evaluate the performance. 
\subsection{Main Results}
We conduct training of SCL on the three aforementioned training datasets and subsequently validate its performance on two large-scale change detection benchmarks. The models are validated on the benchmarks without any additional training to verify the method's generalization. Sequentially, the results are shown in Tab.~\ref{tab:benchmarks}. Tab.~\ref{tab:benchmarks} demonstrates that SCL demonstrates robust generalization and competitive accuracy without task-specific supervised fine-tuning on target datasets LEVIR-CD~\cite{levir-cd} and WHU-CD~\cite{WHU}. This underscores the ability of SCL to generalize well to diverse datasets, showcasing its potential for robust change detection applications. 

In horizontal comparison in Tab.~\ref{tab:benchmarks}, methods are improved  that trained in the SECOND than those trained in xView2 datasets. It reveals that real-world variations present in SECOND contribute to improved generalization for BiT~\cite{BIT}, TinyCD~\cite{codegoni2022tinycd}. However, due to its smaller data scale, performance is notably poor for Changer~\cite{arxiv:changer} and ChangeStar~\cite{Zheng_2021_ICCV_changestar}. In contrast, leveraging our training strategy results in enhanced performance across all methods. It illustrates the validity of our generation method SAIN. In a vertical comparison, our method consistently surpasses competing approaches in the evaluated benchmarks, revealing that many existing methods are prone to training overfitting and exhibit limited generalization and zero-shot performance. Our method has excellent generalization and cross-domain learning abilities. Specifically, when validating on the LEVIR-CD dataset trained on the xView2 dataset, our method outperforms the state-of-the-art ChangeStar~\cite{Zheng_2021_ICCV_changestar} by 9.71\% in F1 and 5.03\% in IoU. Furthermore, when training on our single-temporal generated dataset, our method achieves a 5.88\% improvement in F1 and a 9.49\% increase in IoU compared to ChangeStar~\cite{Zheng_2021_ICCV_changestar} on LEVIR-CD.

\begin{table}[tb]
\centering
    \resizebox{\linewidth}{!}{
    \begin{tabular}{c|c|ccc}
    \toprule
    \multirow{2}*{Model} & \multirow{2}*{backbone} & \multicolumn{3}{c}{LEVIR-CD} \\
    \cline{3-5}
    ~   & ~ & F1 & IoU & OA \\
    \hline
    ChangeStar~\cite{Zheng_2021_ICCV_changestar} &  ResNet-50 & 45.88 & 29.77 & 93.57 \\
    ChangeCLIP~\cite{DONG202453ChangeCLIP} & ViT-B& 59.68 & 42.53 & 96.07 \\
    Changen~\cite{zheng2023changen}  & MiT-B1 & 62.30 & 45.24 & 96.26 \\
    AnyChange~\cite{zheng2024anychange}  & ViT-B & 73.27 & 58.10 & 97.74 \\
    AdaptOVCD~\cite{dou2026AdaptOVCD} & ViT-H & 68.00 & 51.52 & - \\
    SCM~\cite{10642429SCM} & -& 62.80 & 53.67 & 88.80 \\
    \hline
    SCL(Ours)  & ViT-B &\textbf{85.19}& \textbf{74.20} & \textbf{98.59} \\
    \bottomrule
    \end{tabular}
    }
    \caption{Quantitative results of change detection with zero-shot and multi-modal vision-language methods on the LEVIR-CD~\cite{levir-cd}.}
    \label{tab:zero-shot}
    \vspace{-1.5em}
\end{table}

\begin{table}[tb]
    \centering
    \resizebox{\linewidth}{!}{
    \begin{tabular}{c|c|cccc|cc}
        \toprule
         & ~ & $lva$ & $pca$ & DTCO  & SAIN & F1 & IoU \\
         \hline
        1 &  ViT  & &  &   &   &  76.24 & 61.61  \\
        2 &  CLIP & &  &  &  &  79.18  &  65.54 \\
        3 &  CLIP+ & \ding{52} &  &  & &  79.50  &  65.98 \\
        4 &  CLIP+ &  & \ding{52} &  &  &  80.37  &  67.18 \\
        5 &  CLIP+ & \ding{52} & \ding{52} &  &  & 81.11 & 68.23 \\
        6 &  CLIP+ &  &  & \ding{52} &  & 82.28 & 69.90 \\
        7 &  CLIP+ &  &  &  & \ding{52} & 80.91 & 67.95 \\
        8 &  CLIP+ & \ding{52} & \ding{52} & \ding{52} &  & \textbf{89.02} & 70.74 \\
        9 &  CLIP+ & \ding{52} &  \ding{52} & \ding{52} & \ding{52} & 85.19 & \textbf{74.20}\\
        \bottomrule
    \end{tabular}
    }
    \caption{Component effectiveness ablation experiment, test on LEVIR-CD. $lva$ is the local patch-visual alignment, $pca$ is the visual-context alignment.}
    \label{tab:ablation}
    \vspace{-2em}
\end{table}
\subsection{Comparison of Zero-Shot Approaches}
Recently, various methods based on multimodal vision-language models for zero-shot learning have emerged in change detection. To effectively observe the performance between these methods and SCL in domain-generalized change detection, we conduct tests of several models on the LEVIR-CD~\cite{levir-cd} datasets. For fairness, all methods are without training on LEVIR-CD. Tab.~\ref{tab:zero-shot} demonstrates that SCL significantly outperforms other methods in generalized change detection with train-free. Notably, the comparison with ChangeCLIP~\cite{DONG202453ChangeCLIP}, another multimodal vision-language model, further confirms effectiveness and superiority of SCL in enhancing downstream task generalization.
\subsection{Ablation Study}
In order to validate the effectiveness of each component, we conduct a series of ablation experiments, as shown in Tab.~\ref{tab:ablation}. Corresponding to SAIN, we use STAR~\cite{Zheng_2021_ICCV_changestar} as the default image pair construction method if without SAIN. It achieves 3.93\% increasing in IoU and 2.94\% improving in F1, comparing with only use image encoder (ViT) of CLIP (row 1). It demonstrates the feasibility of text-visual contrastive learning. From rows 3-9, we validate the effectiveness of local patch-visual alignment $lva$, visual-context alignment $pca$, DTCO, and SAIN. Comparing row 8 with row 2, we evaluate the combined effect of the alignments and DTCO, which results 9.73\% increase in F1 and 5.2\% increase in IoU. Comparing the row 8 with row 9, SCL exhibits a lower F1 but a higher IoU. This is because SAIN fully simulates changes in natural scenarios, unlike the random matching approach in STAR, which tends to cause excessive sensitivity to changes. It indicates that the SAIN is more effective for domain generalization learning.

\begin{figure*}[tb]
    \centering
    \includegraphics[width=0.8\linewidth]{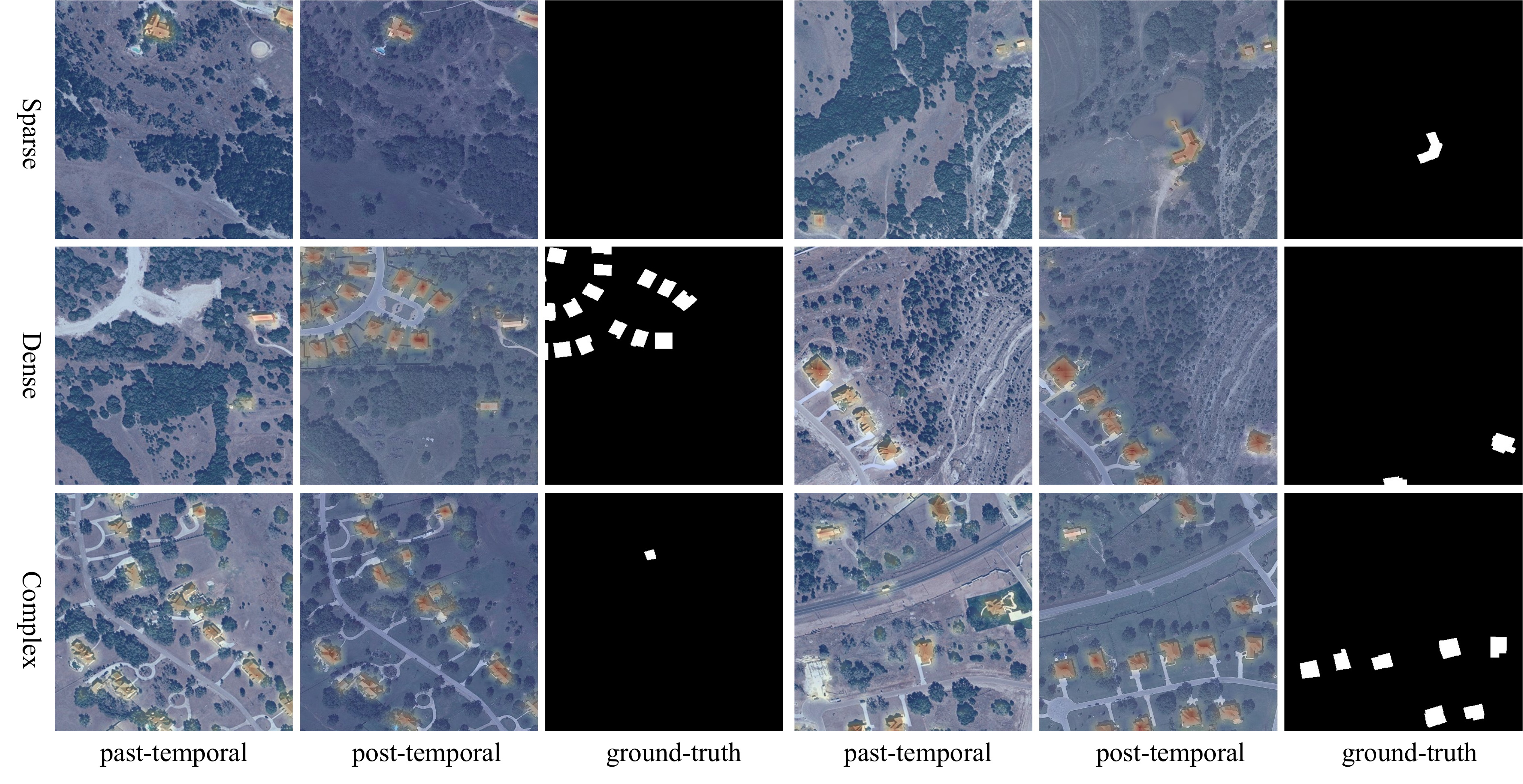}
    \vspace{-2mm}
    \caption{Visualization of SCL activation patterns in different scenes.}
    \label{fig:heatmap}
    \vspace{-1mm}
\end{figure*}

Additionally, we conduct ablation experiments on the hyperparameter in Eq.~\ref{eq:loss-overall}, as shown in Tab.~\ref{tab:hypepara}. Setting $\alpha$ and $\beta$ to 0.1 yields the optimal performance under our experimental settings. It also supports the earlier theoretical proposition of using $\mathcal{L}_{lva}, \mathcal{L}_{pca,po}$ and $\mathcal{L}_{pca,pa}$ as an auxiliary loss in Eq.~\ref{eq:loss-overall}.

To validate the effect of fine-tuning the CLIP~\cite{radford2021learning} image encoder, we conduct comparative experiments on the effect of the learning rate multiplier for CLIP image encoder, as shown in the Tab.~\ref{tab:weight}. We validate on the LEVIR-CD~\cite{levir-cd} dataset. From the Tab.~\ref{tab:weight}, it can be observed that the model achieves stable and favorable performance when it is set to $0.1*lr$ under our experimental setup.
\begin{table}[tb]
\setlength{\tabcolsep}{3pt}
    \centering
    \begin{tabular}[width=\linewidth]{c|ccccc}
        \toprule
        $\alpha, \beta$ &0.1/0.1 &0.1/1.0 &1.0/0.1&0.5/0.5 &1.0/1.0     \\
        \hline
        F1              & 85.19  & 83.58  & 83.24 &  84.71 & 82.84       \\
        IoU             & 74.20  & 71.79  & 71.30 &  73.48 & 70.70      \\
        \bottomrule
    \end{tabular}
    \caption{Ablation study of $\alpha$ and $\beta$ in Eq.~\ref{eq:loss-overall}.}
    \label{tab:hypepara}
    \vspace{-3mm}
\end{table}
\begin{table}[tb]
    \centering
    \setlength{\tabcolsep}{3pt}
    \begin{tabular}{c|ccccccc}
    \toprule
         $lr*$ & 0.01 & 0.02 & 0.05 & 0.07 & 0.09 & 0.10 &0.15\\
         \hline
         F1       &81.87 &82.01 &84.10 &83.06 &84.79 &\textbf{85.19} &83.21  \\
         IoU      &69.31 &69.51 &72.56 &71.03 &73.59 &\textbf{74.20} &71.25  \\
    \bottomrule
    \end{tabular}
    \caption{Ablation experiments with different learning rate multipliers for fine-tuning the CLIP image encoder, where $lr$ denotes the model learning rate.}
    \label{tab:weight}
    \vspace{-1.5em}
\end{table}
\subsection{Visualization Analysis of Change Detection}
To illustrate the change detection performance of SCL, we conduct a visual analysis of the detection results on the LEVIR-CD dataset. As depicted in Fig.~\ref{fig:badcase}, instances of false positives (FP) are denoted in red, and false negatives (FN) are represented in blue. The first row showcases the excellent performance of our method in detecting sparse, small-scale targets. The second row reflects the superior performance of our approach involving both large-scale and small-scale objects, surpassing the capabilities of SCL. The third row demonstrate the outstanding performance of SCL in detecting dense objects. This visual analysis provides a comprehensive representation of SCL's robustness, performing excellently in various scenarios.

We conduct a visual analysis of the activation maps of the bi-temporal segmentation head, as shown in Fig.~\ref{fig:heatmap}. It can be observed that our method SCL focuses solely on discriminative features. And it exhibits excellent robustness in distinguishing between sparse (row 1) and dense objects (row 2), as well as in complex scenes (row 3). There is extensive seasonal change noise in the first row and the second row. SCL detects buildings very accurately for each picture in the bi-temporal image pairs, as shown in Fig~\ref{fig:heatmap}, the red areas in each figure are focused on buildings. Benefiting from that, the change detection is very accurate. The noise on the third row consists of buildings variations in the different season states and wasteland with similar shapes as buildings. The activation maps show that SCL detects changes and effectively learns the building category in bi-temporal images, accurately detecting building categories and exhibiting excellent noise resistance.
\section{Conclusion}
In this paper, we propose a multimodal unified change detection architecture SCL to address the issue of poor generalization in existing change detection methods. To enhance the generalization of text-vision pretraining contrastive learning, we propose a dynamic context optimization method DTCO to improve the model generalization ability. Additionally, we introduce a controllable generation and single-temporal training strategy SAIN for change detection to overcome the reliance of existing methods on large-scale, high-quality annotated image pairs. Without additional training on target dataset, our network architecture exhibits outstanding domain generalization performance on target change datasets. Extensive experiments demonstrate that our approach is effective for single-temporal supervised training and offers insights into the application of unified models in change detection.
{
    \small
    \bibliographystyle{ieeenat_fullname}
    \bibliography{main}
}


\end{document}